\documentclass[journal]{IEEEtran}
\ifCLASSINFOpdf
\else
\fi

\usepackage{amsmath,amsfonts}
\usepackage{algorithmic}
\usepackage{array}
\usepackage[caption=false,font=normalsize,labelfont=sf,textfont=sf]{subfig}
\usepackage{textcomp}
\usepackage{stfloats}
\usepackage{url}
\usepackage{verbatim}
\usepackage{graphicx}
\usepackage{cite}
\usepackage{amsmath}

\usepackage{tabularx}
\usepackage{booktabs}
\usepackage{multirow}

\usepackage{xcolor}

\usepackage[colorlinks,
            linkcolor=black,
            anchorcolor=blue,
            citecolor=green,
            urlcolor=black
            ]{hyperref}

\begin{document}
%
\title{Intra- and Cross-Scale Feature Interaction Network for Referring Remote Sensing Image Segmentation
}
%
%
%

\author{Zhigang Yang, Huiguang Yao, Linmao Tian,
Xuezhi Zhao, Qiang Li,~\IEEEmembership{Member,~IEEE}, 
Qi Wang,~\IEEEmembership{Senior Member,~IEEE}
\thanks{ 
This work was supported by the National Natural Science Foundation of
China under Grant 62471394, Grant 62571437. Zhigang Yang, Qiang Li, and Qi Wang are with the School of Artificial Intelligence, Optics and Electronics (iOPEN), Northwestern Polytechnical University, Xi’an 710072, P.R. China. (e-mail: zgyang@mail.nwpu.edu.cn, liqmges@gmail.com, 
crabwq@gmail.com) (Corresponding authors: Qi Wang.)

Huiguang Yao, Xuezhi Zhao are with School of Computer Science, Northwestern Polytechnical University, Xi’an 710072, P.R. China. (e-mail: yhg2655@mail.nwpu.edu.cn, xuezhizhao@mail.nwpu.edu.cn)

Linmao Tian is with the School of Software, Northwestern Polytechnical University, Xi’an 710072, P.R. China. (e-mail: tianlm0808@mail.nwpu.edu.cn)
}}

%
%

\markboth{IEEE TRANSACTIONS ON Geoscience and Remote Sensing}%
{Shell \MakeLowercase{\textit{et al.}}: Bare Demo of IEEEtran.cls for Journals}
%



\maketitle
\begin{abstract}
Referring remote sensing image segmentation (RRSIS) is a challenging multimodal task that requires accurate localization and pixel-level segmentation of targets specified by natural-language expressions. Existing RRSIS datasets remain limited in image resolution, scene diversity, and category coverage, which restricts the generalization ability and practical applicability of current methods. To advance this field, we introduce NWPU-Refer, the largest and most diverse bilingual RRSIS dataset to date. It contains 15,003 high-resolution images (1024--2048 px) collected from more than 30 countries across five continents, together with 49,745 annotated targets covering single-object, multi-object, and no-target scenarios. We further propose the Multi-scale Referring Segmentation Network (MRSNet), a framework tailored to the characteristics of remote sensing imagery. Specifically, MRSNet incorporates an Intra-scale Feature Interaction Module (IFIM) to refine fine-grained features within each encoder stage and a Hierarchical Feature Interaction Module (HFIM) to integrate cross-scale information while preserving spatial structure. Extensive experiments on NWPU-Refer and RRSIS-D demonstrate that MRSNet achieves strong and competitive performance across multiple metrics, especially under strict IoU thresholds and IoU-based evaluation, validating the effectiveness of the proposed dataset and model. The dataset and code are publicly available at \textcolor[RGB]{237,2,140}{https://github.com/CVer-Yang/NWPU-Refer}.

\end{abstract}


%
\IEEEpeerreviewmaketitle

\section{Introduction}
\IEEEPARstart{R}{eferring} segmentation,  as a pioneering task at the intersection of computer vision \cite{wei2024stronger,11506600} and natural language processing \cite{zhi2024lscenellm,di2023retrieval}, has garnered significant attention due to its ability to segment specific target regions based on user-provided natural language descriptions \cite{huang2022scaleformer, liu2023gres, shang2024prompt, chng2024mask,ding2023mevis,karazija2021clevrtex,chen2020scanrefer}. This task is distinguished by its interactivity and flexibility, enabling models to interpret user intent and generate personalized segmentation results. However, the unique characteristics of remote sensing images \cite{li2026spatial,li2024edge,yang2026semantic}, including their diverse perspectives, varying scales, and intricate relationships, pose substantial challenges to existing RRSIS methods.

\begin{figure}[tb]
\centering
\includegraphics[width=3.4in]{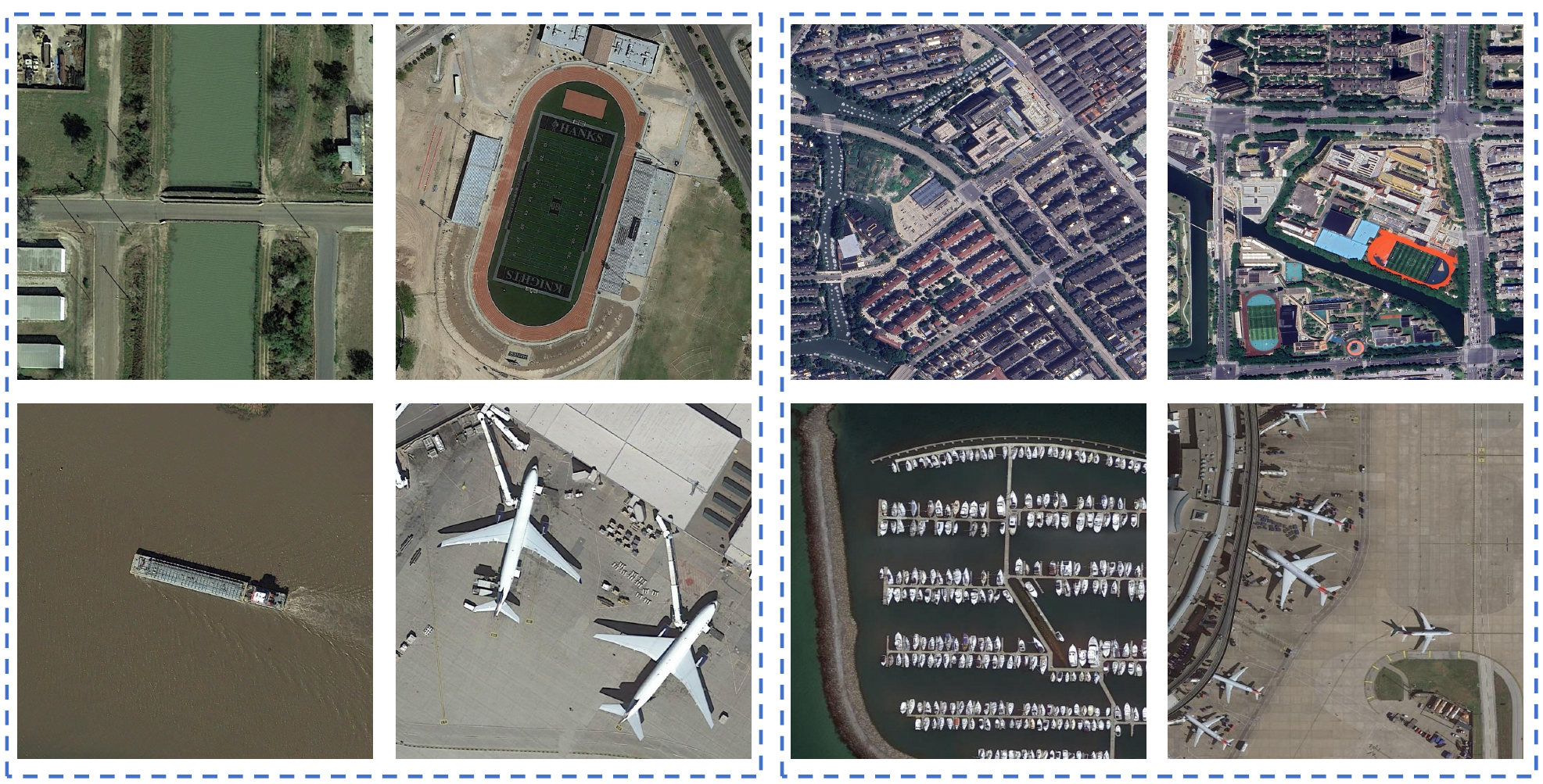}
\vspace{1pt}
\caption{Comparison of refer segmentation datasets for four target categories: bridges, sports fields, ships, and airplanes. As shown in the left column, existing datasets primarily consist of simple scenes with prominent and easily segmented objects. In contrast, our proposed dataset (right column) introduces greater detail and complexity, better capturing the intricate real-world environments encountered in practical applications.}
\label{fig_1}
\end{figure}
While the referring segmentation task has garnered significant attention in recent years, existing datasets \cite{yuan2024rrsis, liu2024rotated} for Referring Remote Sensing Image Segmentation remain constrained by notable limitations. These datasets are often derived through simplistic secondary processing of existing object detection datasets, which is insufficient to meet the complex and diverse requirements of real-world applications. Consequently, this results in suboptimal data quality and limited diversity. On one hand, annotations in these datasets are frequently generated using automated tools such as the Segment Anything Model (SAM) \cite{kirillov2023segment}.   Notwithstanding their efficiency, these frameworks are prone to introducing subtle inaccuracies and omitting fine-grained details, which are critical for precise segmentation tasks. 
On the other hand, as illustrated in Fig. \ref{fig_1}, existing datasets predominantly focus on simplified scenarios with salient, isolated targets (e.g., a single ship or a sports field), failing to capture the complexity of real-world remote sensing images. These real-world images typically feature intricate backgrounds and multi-object contexts, which pose significant challenges for accurate segmentation. Furthermore, the relatively outdated nature of the imagery gradually compromises the generalization capabilities of models, thereby undermining their performance in practical applications.

To alleviate these challenges, we address RRSIS from a single design perspective: intra-scale refinement before hierarchical cross-scale integration. The motivation is that remote sensing refer segmentation first requires reliable local discrimination within each encoder stage, especially for small objects and relation-sensitive queries, and then needs structure-preserving aggregation across stages so that the refined semantics are not diluted during fusion. Guided by this principle, we conduct a comprehensive large-scale data collection effort across five continents and build NWPU-Refer, which captures diverse urban, rural, forested, desert, and oceanic scenes. We further extend the task definition by introducing category-based referring expressions, such as ``all vehicles,'' ``the first row of photovoltaic panels,'' and ``vehicles traveling left on the road,'' while also supporting no-target scenarios. In addition, we design MRSNet, where IFIM performs stage-wise refinement of multi-scale details, spatial relations, and cross-modal alignment, and HFIM subsequently performs hierarchical cross-scale aggregation without discarding scale-specific semantics. The principal contributions of this work are summarized as follows:

\begin{itemize}

\item We present NWPU-Refer, a global-scale bilingual referring remote sensing image segmentation dataset covering diverse geographic and environmental scenarios across five continents. The dataset captures the complexity of real-world applications and introduces multi-target and no-target settings, thereby broadening the scope of RRSIS.

\item We propose MRSNet, which is built on the principle of intra-scale refinement before hierarchical cross-scale integration. IFIM enhances each encoder stage with pyramidal spatial-spectral refinement, local relation modeling, and cross-modal alignment, while HFIM aggregates the refined stage features through shared-resolution fusion and inverse restoration.

\item Experiments are conducted on NWPU-Refer dataset to evaluate the proposed method across a wide range of scenarios. The results demonstrate substantial performance improvements compared to existing approaches, confirming the effectiveness of the proposed work.
\end{itemize}

\section{Related Work}
In this section, we review the relevant datasets and refer segmentation methods.
\subsection{Reference Image Segmentation  Datasets}

The development of referring segmentation has been significantly propelled by numerous high-quality datasets \cite{li2024transformer,wang2024hierarchical}. Among these, the RefCOCO dataset \cite{kazemzadeh2014referitgame}, an extension of the MSCOCO dataset \cite{lin2014microsoft}, provides annotations specifically designed for referring segmentation. This dataset facilitates the exploration of the intricate relationships between language descriptions and visual targets by integrating language and vision modalities. Building upon this foundation, the gRefCOCO dataset \cite{liu2023gres} extends the scope of referring expressions to include descriptions referencing multiple targets within a single image, thereby enhancing its representativeness for real-world applications.
In the domain of remote sensing imagery, the RSIS dataset stands as the first benchmark explicitly developed for referring segmentation. It leverages the Segment Anything Model (SAM) \cite{kirillov2023segment} to generate segmentation masks from the RSVGD dataset \cite{zhan2023rsvg}, paired with corresponding natural language descriptions. Despite these advancements, existing remote sensing datasets exhibit notable limitations. These datasets predominantly focus on salient objects, limiting their ability to address challenging scenarios involving small or low-contrast targets. Furthermore, the image sources lack temporal and spatial diversity, often being restricted to outdated scenes with limited complexity. Additionally, the annotations are primarily in English, which hinders their applicability in multilingual regions and reduces their relevance to diverse real-world applications.
\subsection{Referring Image Segmentation Methods}

Building on the foundations laid by these datasets, researchers \cite{huang2020referring,hu2020bi,liu2021cross,lei2024exploring,ha2024finding,yang2024remamber} have proposed various methods for language-guided \cite{11062613} referring segmentation \cite{wang2025iterprime,huang2025densely,nguyen2025vision,yue2024adaptive} and reasoning segmentation \cite{lai2024lisa,ren2024pixellm,yan2025visa}. Early approaches \cite{li2018referring,jing2021locate,liu2023referring} primarily rely on convolutional neural network (CNN) based architectures, where multimodal fusion is achieved through simple concatenation of language embeddings and image features. For instance, the CMSA model \cite{ye2019cross} introduces cross-modal self-attention mechanisms to compute similarities between language descriptions and image regions, generating initial segmentation results. However, these methods often struggle with limited fusion capabilities, making it challenging to capture the complex contextual information embedded in language descriptions.

Transformer-based approaches \cite{vaswani2017attention} have brought significant advancements to this field. For example, the Language-Aware Vision Transformer (LAVT) \cite{yang2022lavt} enables deep interaction between language embeddings and image features, effectively capturing fine-grained semantics in language descriptions. This improves the understanding of complex linguistic instructions and enhances segmentation accuracy. Similarly, the RMSIN model integrates multiple visual and linguistic gating mechanisms to align features, while its adaptive rotational convolutions in the decoder ensure precise segmentation of referred objects.
Despite these advancements, substantial challenges remain. Remote sensing images encompass a wide range of object scales, from large targets such as ground tracks to small objects like vehicles that occupy only a few pixels. This necessitates models that can effectively perceive and represent features across multiple scales. Furthermore, remote sensing referring segmentation tasks demand enhanced positional sensitivity, requiring models to accurately capture spatial relationships, including vertical and horizontal distinctions. Additionally, certain scenarios involve the absence of target objects, necessitating models to selectively align visual and linguistic features to achieve robust segmentation. Cross-scale feature aggregation \cite{yang2024c,yang2024hcnet} has demonstrated substantial benefits in segmentation tasks by enabling effective feature modeling and integration. Incorporating cross-scale feature aggregation into remote sensing referring segmentation offers a promising direction for improving feature representation and segmentation performance. This approach not only addresses the challenges posed by multi-scale objects but also enhances the model's ability to handle complex spatial relationships and diverse scenarios.

\section{NWPU-Refer Dataset}
This section describes NWPU-Refer in terms of its annotation protocol, dataset statistics, representative scenarios, and comparison with existing remote sensing referring segmentation benchmarks.

\subsection{Dataset Annotation}

To ensure precise and high-quality data annotations, a systematic process is designed with three key components.

\textit{Step 1: Target category definition.} A comprehensive set of 32 target categories commonly observed in remote sensing imagery is defined to ensure diverse and structured representation. These categories include transportation targets such as cars, ships, trains, and airplanes; infrastructure including bridges, roads, road intersections, buildings, and airport runways; natural and artificial scenes such as lakes, rivers, grasslands, open areas, and oceans; sports facilities including basketball courts, ground track fields, soccer fields, and tennis courts; industrial equipment such as wind turbines, power line towers, storage tanks, and construction towers; and other objects including parking lots, dams, chimneys, and containers.

\textit{Step 2: Multi-language Annotation.} To support cross-lingual referring segmentation, bilingual expressions are created within the annotation protocol for the predefined 32 target categories. The annotation process covers six dimensions: size-based references, spatial relationships, color attributes, category relations, motion states, and associative relationships. The Chinese and English expressions are written to preserve the same target semantics while remaining natural in each language.

\textit{Step 3: Data Labeling and Quality Assurance.} 
A team of four experts conducts the annotation process. Three trained annotators label target categories, attributes, and spatial relationships, while a senior annotator reviews completeness, accuracy, bilingual semantic consistency, and mask-expression alignment. Particular attention is paid to spatial phrases such as left/right, top/bottom, and relational attributes so that the final expressions remain consistent with the annotated regions. All annotations are manually performed without auxiliary tools, ensuring high precision and reliability.
\begin{figure*}[ht]
\centering
\includegraphics[width=7.0in]{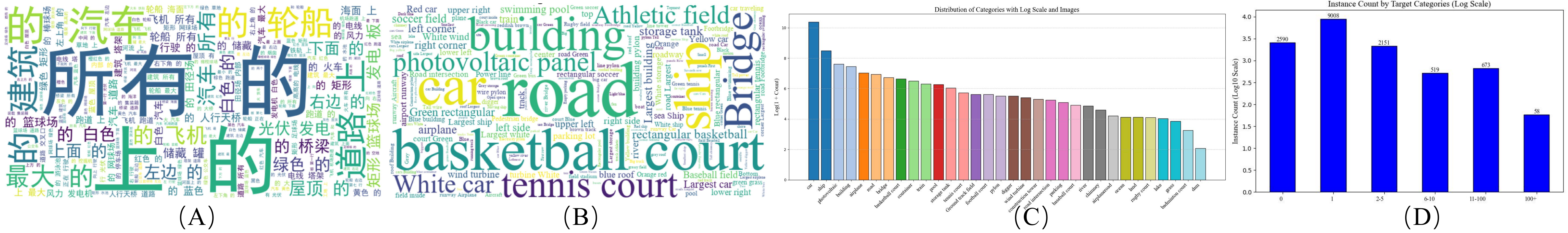}
\caption{Illustration of data distribution characteristics. (A)--(B) Word clouds from annotations highlight the frequency of target-related terms. Common targets (e.g., cars and ships) appear more frequently, while rare targets (e.g., badminton courts) appear less often. (C) The annotation-category distribution reflects the class imbalance between frequent and rare targets. (D) The annotation-type distribution shows the proportion of single-target, multi-target, and non-target descriptions.}
\label{fig_4}
\end{figure*}
\subsection{Dataset Analysis}

\begin{figure*}[ht]
\centering
\includegraphics[width=7.0in]{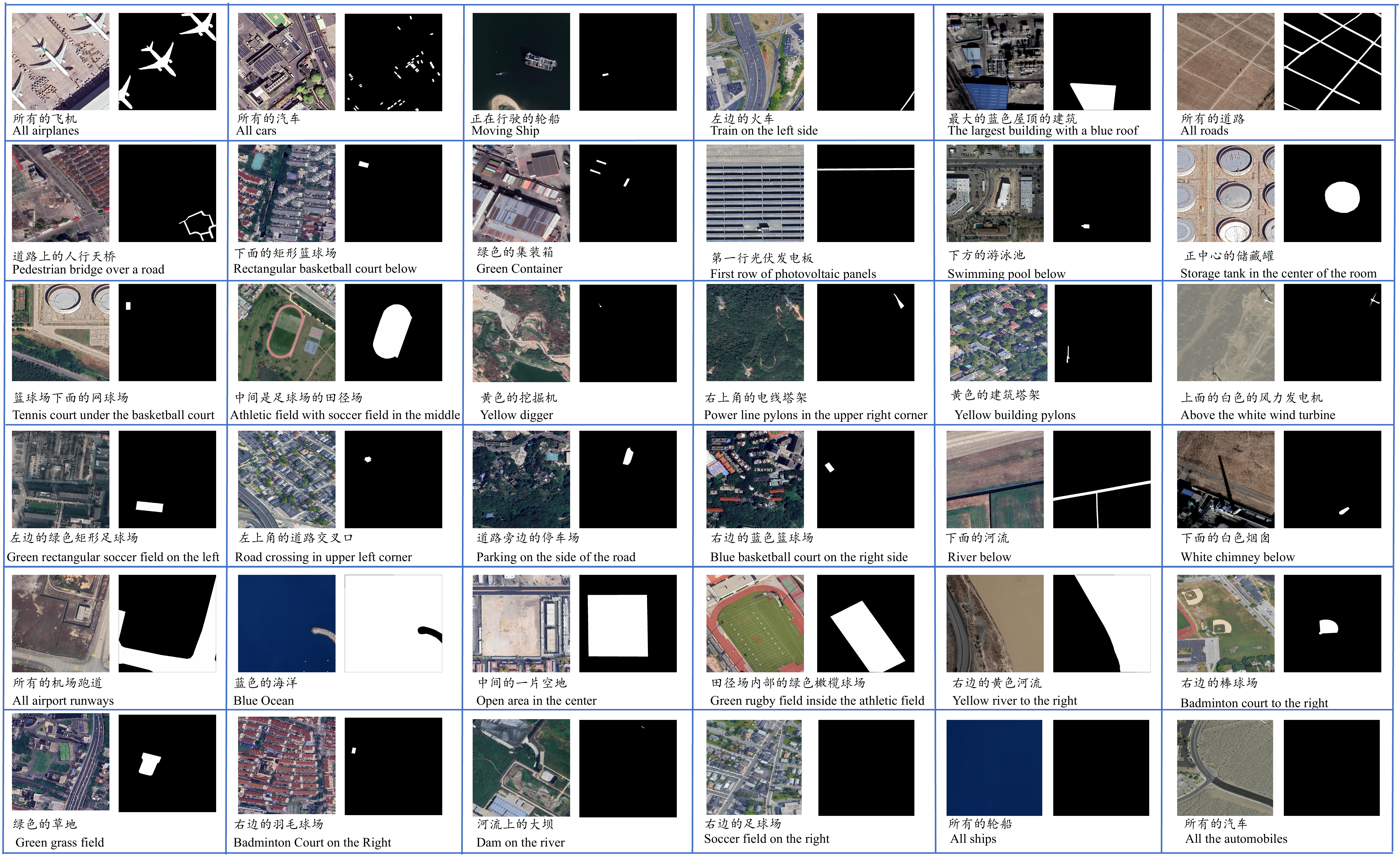}
\vspace{1pt}
\caption{Visualization examples of diverse segmentation tasks in the NWPU-Refer dataset. Each sample includes Chinese and English expressions, together with the corresponding target regions.}
\label{fig_5}
\end{figure*}
The NWPU-Refer dataset is a comprehensive benchmark for remote sensing image segmentation, containing high-resolution imagery (0.12–0.5 m) covering diverse scenarios.
Fig. \ref{fig_4}(A) and (B) present word clouds of dataset annotations, highlighting that common targets including cars, ships, and airplanes dominate the instances, while rare targets such as badminton courts, dams, and golf courses account for a small proportion. This category imbalance, further illustrated in Fig. \ref{fig_4}(C), presents significant challenges for model training and generalization across all categories.
Additionally, Fig. \ref{fig_4}(D) shows the distribution of annotation types: single-target descriptions are most prevalent, followed by multi-target descriptions, with target-free descriptions being the least common. This indicates that models should be optimized for efficient single-target processing while maintaining capability for complex multi-target scenarios.
\subsection{Typical Scene Analysis}
To intuitively demonstrate the diversity of the dataset, we conduct a visual analysis of each annotated object category. As shown in Fig. \ref{fig_5}, the targets in the dataset exhibit significant diversity and complexity, including buildings of various shapes, vehicles of different sizes, and vehicle categories with intricate relationships. In addition, we analyze special cases in the dataset, such as occluded targets and blurry target scenarios. In these cases, the boundaries of the targets are ambiguous, or their semantics are unclear, posing greater challenges to the segmentation accuracy of models. However, these challenging samples also provide valuable opportunities for evaluating model performance in complex scenarios.
\subsection{Comparison with Existing Datasets}

\begin{table*}[htb]
    \caption{Comparison with existing remote sensing referring segmentation datasets in terms of spatial resolution, image size, annotation scale, object-category coverage, attribute diversity, and annotation generation.}
    \centering
    \small
    \setlength{\tabcolsep}{3pt} 
    \renewcommand\arraystretch{1.1} 
    \begin{tabular}{m{1.8cm}<{\centering}|m{2.0cm}<{\centering}|m{1.2cm}<{\centering}|m{1.5cm}<{\centering}|m{1.4cm}<{\centering}|m{1.4cm}<{\centering}|m{1.4cm}<{\centering}|m{1.4cm}<{\centering}|m{1.4cm}<{\centering}|m{1.8cm}<{\centering}}
    \hline
    \hline
    Dataset & Image Resolution / Size & Images & Annotations & Object Categories & Attribute Types & Single-object & Multi-object & Non-object & Annotation Generation \\ \hline
    RefSegRS & 0.13m / 512 & 4420 & 4420 & 20 & 3 & $\checkmark$ & $\checkmark$ & $\times$ & Manual \\ \hline
    RRSIS-D & 0.5m--30m / 800 & 17402 & 17402 & 20 & 7 & $\checkmark$ & $\times$ & $\times$ & Semi-auto \\ \hline
    NWPU-Refer & 0.12m--0.5m / 1024-2048 & 15003 & 49745 & 32 & 6 & $\checkmark$ & $\checkmark$ & $\checkmark$ & Manual \\
    \hline
    \hline
    \end{tabular}
    \label{tab1}
\end{table*}

Table \ref{tab1} compares three remote sensing referring segmentation datasets: RefSegRS, RRSIS-D, and NWPU-Refer. RefSegRS contains 4,420 image-text-mask triplets, 20 object categories, and 3 attribute types. RRSIS-D includes 17,402 triplets, 20 object categories, and 7 attribute types, but relies on semi-automatic annotation. In contrast, NWPU-Refer comprises 15,003 images with 49,745 manually generated annotations, 32 object categories, and 6 annotation dimensions, while also supporting single-object, multi-object, and non-object expressions. This broader semantic coverage, together with manual verification, makes NWPU-Refer a more comprehensive benchmark for complex remote sensing referring segmentation.

\section{Methods}
\subsection{Overview}
\begin{figure*}[ht]
\centering
\includegraphics[width=6.2in]{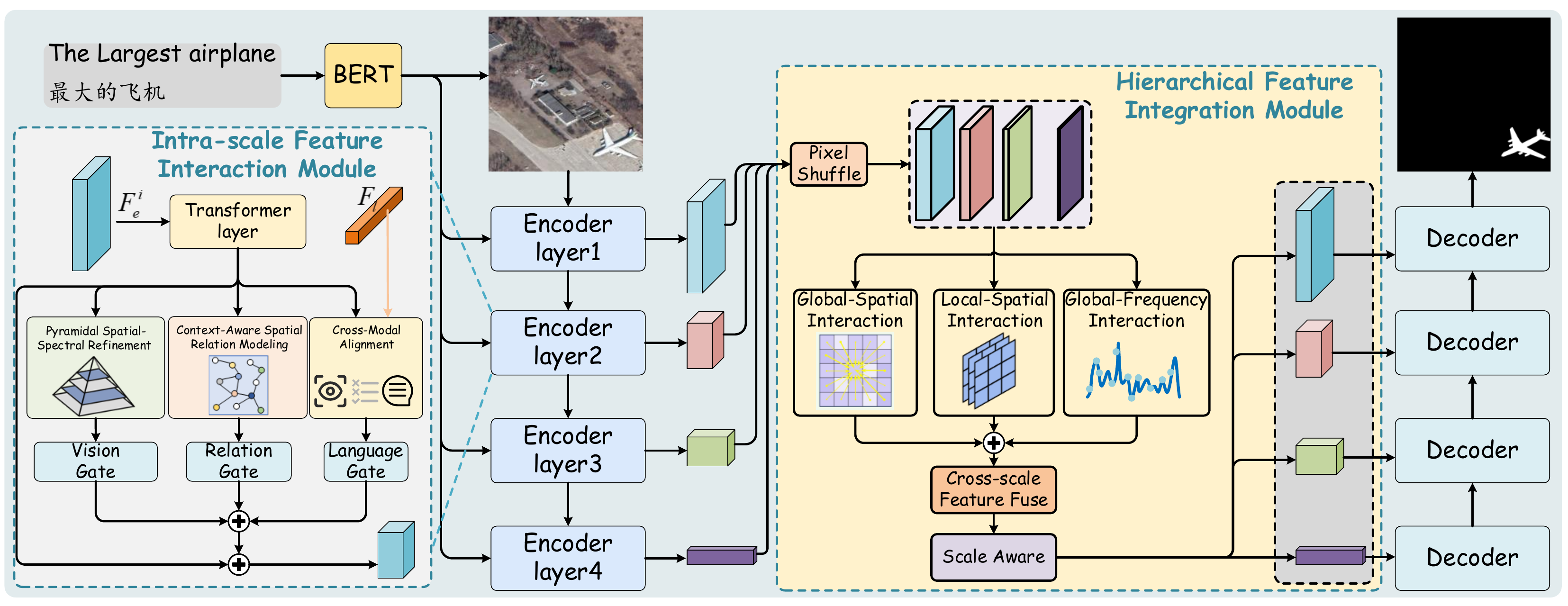}
\vspace{1pt}
\caption{Overview of the proposed MRSNet. The inputs are a remote sensing image and language tokens. Encoder layers denote the four stages of the Swin-Base backbone, and each stage is equipped with an Intra-scale Feature Interaction Module (IFIM). The resulting stage features are passed to the Hierarchical Feature Integration Module (HFIM) for cross-scale aggregation before decoder prediction.}
\label{fig_6}
\end{figure*}

The architecture of MRSNet is depicted in Fig. \ref{fig_6}. It comprises IFIM inserted into each of the four encoder stages, HFIM for hierarchical cross-scale integration, and a decoder for final mask prediction. We instantiate the visual encoder with Swin-Transformer \cite{liu2021swin} and the language encoder with BERT \cite{devlin2018bert}.

Given a remote sensing image \( I \in \mathbb{R}^{H \times W \times 3} \) and language tokens \( L \in \mathbb{R}^{N \times C} \), MRSNet first refines stage-wise features \(\mathbf{F}_{1}, \mathbf{F}_{2}, \mathbf{F}_{3}, \mathbf{F}_{4}\) through IFIM so that each stage explicitly models intra-scale semantics, local spatial relations, and cross-modal alignment. HFIM then maps the refined stage features to a shared resolution, fuses local, global, and frequency cues, splits the fused representation according to the original branch sizes, and restores each stage through inverse PixelUnshuffle operations. Finally, the decoder generates the segmentation mask associated with the input expression.

\subsection{Intra-Scale Feature Interaction Module}

The Intra-Scale Feature Interaction Module is designed to perform three key functions: capturing multi-scale visual features, modeling contextual relationships between targets, and achieving visual-linguistic alignment. This module enhances feature completeness and discriminability, significantly advancing complex scene referring segmentation capabilities.

\begin{figure}[ht]
\centering
\includegraphics[width=3.5in]{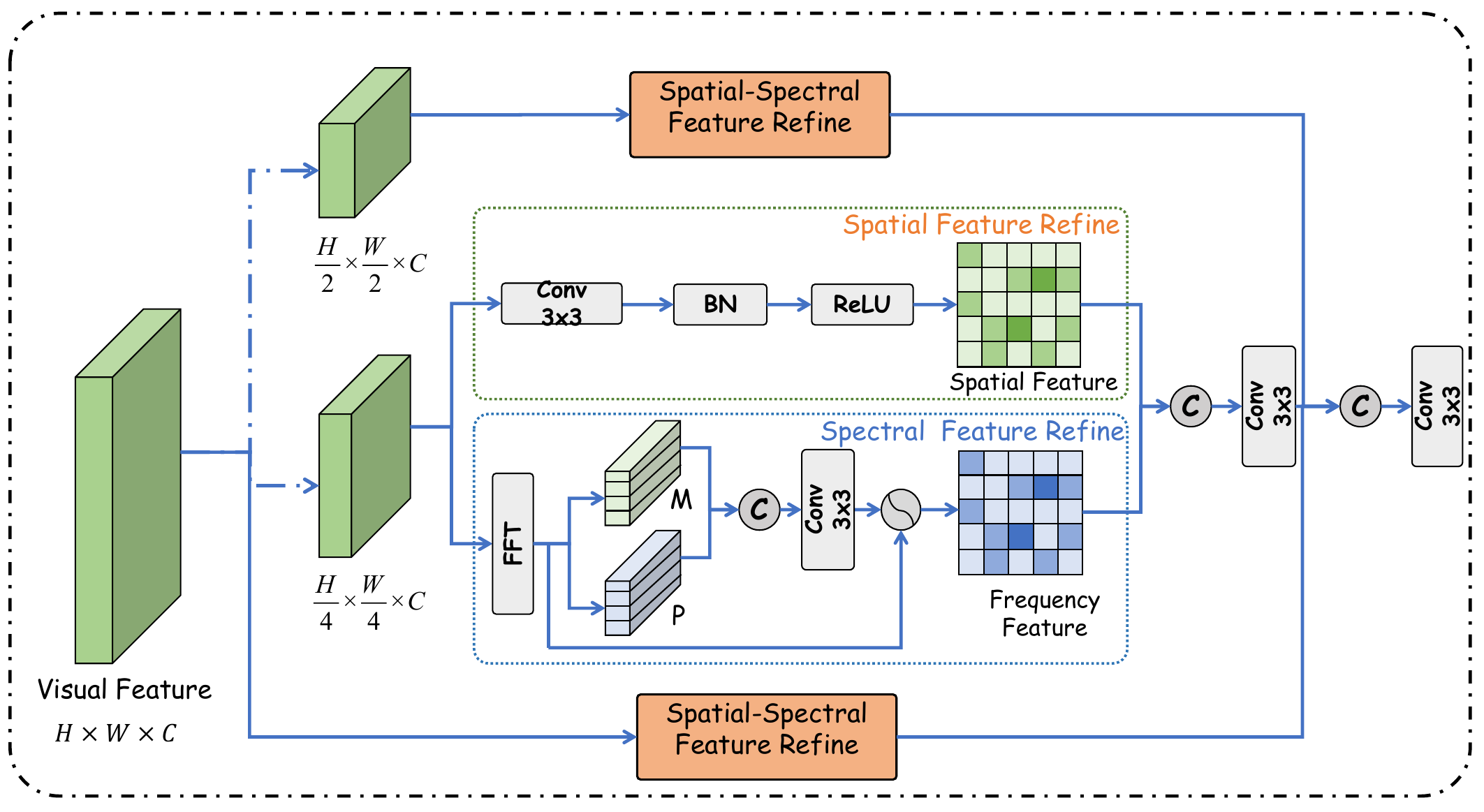}
\caption{The structure of the proposed pyramidal spatial-spectral refinement submodule.}
\label{fig_71}
\end{figure}

\paragraph{Pyramidal Spatial-Spectral Refinement Submodule} As shown in Fig. \ref{fig_71}, the visual feature map is refined in both spatial and frequency domains across multiple scales, generating comprehensive feature representations with robust intra-scale perception. This is particularly important for handling objects of different scales in remote sensing images.

\textit{Spatial-Spectral Refinement:} Given the input feature map \({X} \in \mathbb{R}^{B \times N \times \text{dim}}\), it is reshaped into a feature map \({X}_{\text{2D}} \in \mathbb{R}^{B \times \text{dim} \times H \times W}\), where \(H \times W = N\). Then, a spatial refined feature \({X}_s\) is generated through a convolution operation. Simultaneously, a Fourier Fast Transform is applied to \({X}_{\text{2D}}\) to extract the frequency domain magnitude \({M}\) and phase \({P}\). These components are concatenated and processed through a convolution layer followed by a softmax activation to generate frequency weights:
\[
{W}_f = \text{Softmax}(\text{Conv}([M, P]))\odot{X}_{\text{2D}},
\] where \(\odot\) denotes element-wise multiplication. The frequency domain features \(X_f\) are then obtained by applying these weights to \(X_{2\text{D}}\). Finally, the spatial features \(X_s\) and frequency domain features \(X_f\) are fused through concatenation followed by a convolution operation to produce the refined feature map. These weights are applied to ${X}_{\text{2D}}$ using element-wise multiplication to generate the frequency domain features ${X}_f$.
Finally, the spatial features ${X}_s$ and the frequency domain features ${X}_f$ are fused through concatenation, followed by a convolution operation.

\textit{Pyramidal Feature Aggregation:}
To capture multi-scale target features, the input feature map is downsampled by factors of 2 and 4, resulting in low-resolution feature maps  \({X}_{\text{2D}}^{1/2}\) and \({X}_{\text{2D}}^{1/4}\), respectively. At each scale, spatial and frequency domain attention features are extracted and fused using the aforementioned \textit{Pyramidal Spatial-Spectral Refinement Submodule} process. The aggregated features are upsampled to the original resolution and concatenated to form the final pyramidal feature map $F_{py}$.

\begin{figure}[ht]
\centering
\includegraphics[width=3.5in]{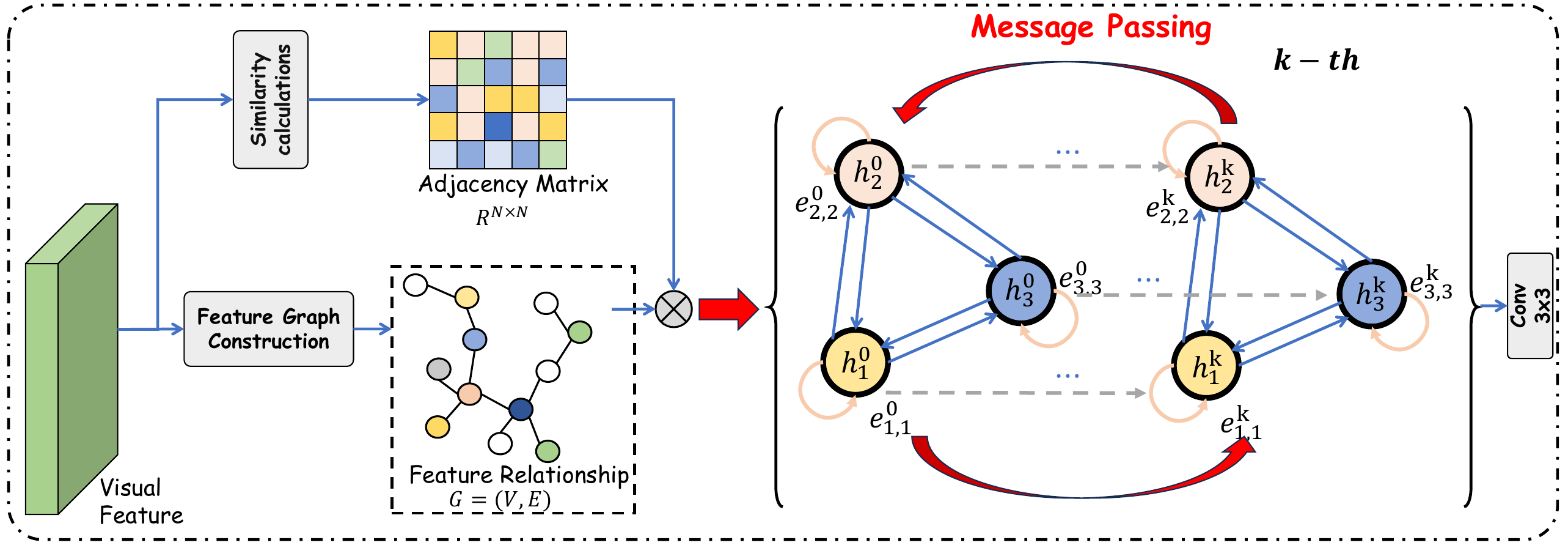}
\caption{The structure of the proposed context-aware spatial relation modeling submodule.}
\label{fig_72}
\end{figure}

\paragraph{Context-Aware Spatial Relation Modeling Submodule} 
For referring segmentation, spatial relations must be modeled explicitly rather than being left to implicit convolutional context alone. As shown in Fig. \ref{fig_72}, the input feature map \(\mathbf{X}_{2D} \in \mathbb{R}^{B \times \text{dim} \times H \times W}\) is flattened into \(\mathbf{X}_{\text{flat}} \in \mathbb{R}^{B \times \text{dim} \times N}\), where \(N = H \times W\), and each spatial position is treated as a node. We define a spatial adjacency matrix \(\mathbf{A} \in \mathbb{R}^{N \times N}\) over a fixed \(3 \times 3\) neighborhood including the center pixel:
\[
\mathbf{A}_{ij} =
\begin{cases}
1, & \text{if } j \text{ lies in the } 3 \times 3 \text{ neighborhood centered at } i; \\
0, & \text{otherwise}.
\end{cases}
\]
We denote the degree matrix as \(\mathbf{D}\), and accordingly construct the row-normalized adjacency matrix \(\tilde{\mathbf{A}} = \mathbf{D}^{-1}\mathbf{A}\). This normalization standardizes the influence of neighboring pixels and keeps the aggregation stable across different locations.

\textit{Feature Aggregation between Points:}
Using \(\tilde{\mathbf{A}}\), the flattened features are aggregated as
\[
\mathbf{C}_{\text{context}} = \mathbf{X}_{\text{flat}}\tilde{\mathbf{A}},
\]
which preserves the feature dimension while propagating local neighborhood context to each node. This operation is especially helpful for relation-sensitive expressions whose interpretation depends on nearby objects or background structures.

\textit{Graph Convolution and Feature Update:}
The contextual features are then updated through a lightweight GCN-style transformation implemented with two learnable \(1 \times 1\) projections:

\[
{F}_{\text{relationship}} = \phi_{2}\!\left(\phi_{1}\!\left(\mathbf{C}_{\text{context}}\right)\right),
\]
where \(\phi_{1}\) and \(\phi_{2}\) denote linear projections. This update captures pixel-wise relationships while keeping the modeling local and computationally tractable.

\begin{figure}[ht]
\centering
\includegraphics[width=3.5in]{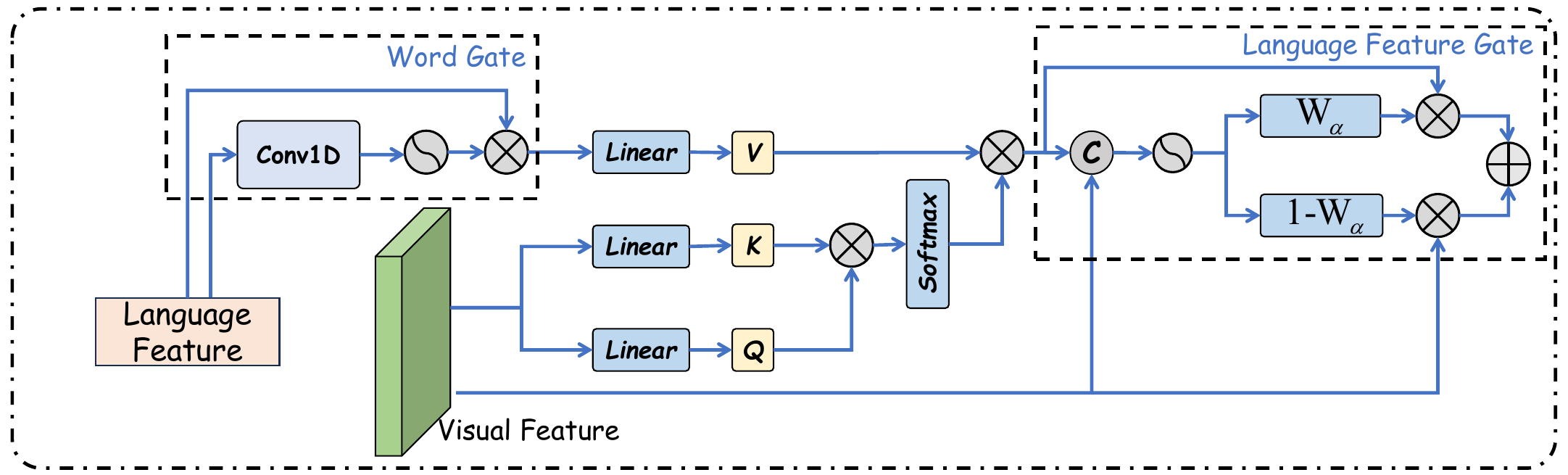}
\caption{The structure of the proposed Cross-Modal Alignment submodule.}
\label{fig_73}
\end{figure}

\paragraph{Cross-Modal Alignment Submodule}
This submodule selectively aligns visual features with the textual query. By incorporating pixel-wise adaptive fusion, it integrates image and language features in a way that is more discriminative for referring segmentation. The structure of this submodule is shown in Fig. \ref{fig_73}.

\textit{Image-Language Interaction:} The input visual feature ${X_{2D}}$ is first projected through a 1D convolution followed by an activation function, producing transformed visual features  $ {V}_{\text{proj}} \in {R}^{B \times \text{dim} \times N}$. To model semantic relationships between visual and linguistic features, the image-language interaction mechanism proceeds as follows: The projected visual features ${X_{\text{proj}}}$ are mapped into queries ${Q}$, while the linguistic features $\mathbf{L} \in \mathbb{R}^{B \times \text{l\_in} \times N_l}$ are mapped into keys ${K}$ and values ${V}$. A dynamic language importance weight $\mathbf{G}_{\text{lang}}$ is calculated to represent the semantic relevance of each word:
    \[
    \mathbf{G}_{\text{lang}} = \sigma(\text{Conv1D}(\mathbf{L})),
    \]
where $\sigma$ denotes the Sigmoid activation function. Subsequently, the input words are weighted by applying these weights to enhance the semantic representation of the linguistic features. Cross-modal relationships are captured by computing the similarity between visual queries ${Q}$ and linguistic keys ${K}$, resulting in an attention map ${S}$. The attention map ${S}$ is then used to weight the linguistic values $\mathbf{V}$, producing the enhanced visual features $V_{L}$ that are aligned with the linguistic features.
The attended linguistic features ${V}_{\text{L}}$ are projected to match the input visual features for subsequent fusion. 

\textit{Cross-Modal Feature Fusion:} 
The projected visual features ${V}_{\text{proj}}$ and the attended visual features ${V}_{\text{L}}$ are concatenated along the channel dimension to form fused features. An attention mechanism is employed to compute the fusion weights ${W}_{\text{proj}}$, which are then applied to combine the these features
:
\[
{F}_{\text{cm}} = {W}_{\text{proj}} \cdot {V}_{\text{proj}} + (1 - {W}_{\text{proj}}) \cdot {V}_{\text{L}}.
\]The fused features are projected through a 1D convolution to produce the final cross-modal feature representation $F_{cm}$.
\paragraph{Adaptive Feature Fusion Module}
The three branch outputs \({F}_{py}\), \({F}_{\text{relationship}}\), and \({F}_{cm}\) are concatenated along the channel dimension. Gated linear projections estimate branch-wise weights, and the weighted features are fused to produce the refined stage output. This design keeps the contributions of pyramidal refinement, spatial relation modeling, and cross-modal alignment explicitly balanced within IFIM.

\subsection{Hierarchical Feature Integration Module}

To address the challenge of integrating multi-scale feature information in complex visual tasks, we propose a Hierarchical Feature Integration Module. Stage-1 features are kept at the reference resolution, while deeper-stage features are mapped to the same resolution through PixelShuffle operators with factors 2, 4, and 8. The aligned features are concatenated to form a shared representation, jointly fused in the spatial and frequency domains, and then restored to their original scales by inverse PixelUnshuffle operations after channel splitting.

\paragraph{Spatial domain feature Integration}

To extract spatial features, the input features are first processed through depthwise separable convolutions, batch normalization, and ReLU activation. This process efficiently captures local neighborhood information, producing local features  \({F}_{\text{local}}\).

To capture global features, we employ a multi-head self-attention mechanism. The input feature map is first projected into three distinct spaces to derive queries, keys, and values. The attention scores are computed by measuring the similarity between queries and keys, followed by a softmax operation to obtain normalized attention weights. These weights are subsequently applied to the values to generate the global features \({F}_{\text{global}}\). 

\paragraph{Frequency Domain Feature Integration}
To capture global frequency information, the input features are transformed into their frequency domain representation \({F}_{\text{freq}}\) using a Fast Fourier Transform. In this domain, normalized similarities between the query \({Q_f}\) and key \({K_f}\) are computed, and the value \({V_f}\) are weighted to generate frequency-domain features:
\[
{F}_{\text{freq}} = \text{IFFT}\left(\text{Softmax}\left(\frac{{Q_f} \cdot {K_f}^\top}{\sqrt{d_k}}\right) \cdot {V_f}\right).
\]
The resulting frequency features are transformed to the spatial domain through an inverse Fourier transform, yielding  \({F}_{\text{freq}} \in \mathbb{R}^{B \times C \times H \times W}\).

\paragraph{Feature Fusion}
The local features \({F}_{\text{local}}\), global features \({F}_{\text{global}}\), and frequency features \({F}_{\text{freq}}\) are concatenated along the channel dimension to form the unified feature \({F}_{\text{fuse}} = [{F}_{\text{local}}; {F}_{\text{global}}; {F}_{\text{freq}}]\). A \(1 \times 1\) convolution is then applied to compress the concatenated representation and produce the final fused feature map.

\paragraph{Hierarchical Feature Division}
After fusion, the shared feature map is split according to the original branch channel sizes so that each group corresponds to one encoder stage. The split groups are then restored through inverse PixelUnshuffle operators, yielding stage-specific outputs with their original resolutions and channel layouts. This split-and-restore procedure enables cross-scale interaction while preserving the semantic identity of each branch.



\begin{table*}[h!]
\caption{Performance comparison across methods on the NWPU-refer dataset. The best results are in bold.}
\centering
\renewcommand{\arraystretch}{1.1} 
\setlength{\tabcolsep}{4pt} 
\begin{tabular}{l|c|c|c|c|c|c|c|c|c|c|c|c|c|c|c}
\hline
\hline
\multirow{2}{*}{\textbf{Method}} & \multirow{2}{*}{\textbf{Publication}} & \multicolumn{2}{c|}{\textbf{P@0.5}} & \multicolumn{2}{c|}{\textbf{P@0.6}} & \multicolumn{2}{c|}{\textbf{P@0.7}} & \multicolumn{2}{c|}{\textbf{P@0.8}} & \multicolumn{2}{c|}{\textbf{P@0.9}} & \multicolumn{2}{c|}{\textbf{oIoU}} & \multicolumn{2}{c}{\textbf{mIoU}} \\
\cline{3-16}
& & \textbf{val} & \textbf{test} & \textbf{val} & \textbf{test} & \textbf{val} & \textbf{test} & \textbf{val} & \textbf{test} & \textbf{val} & \textbf{test} & \textbf{val} & \textbf{test} & \textbf{val} & \textbf{test} \\
\hline
LAVT \cite{yang2022lavt} & \textit{CVPR 22} & 37.46 & 37.04 & 31.78 & 30.99 & 25.22 & 23.29 & 17.16 & 15.50 & 7.31 & 6.92 & 62.23 & 55.61 & 34.66 & 33.90 \\
RMSIN \cite{liu2024rotated}& \textit{CVPR 24} & 44.57 & \textbf{47.95} & 38.69 & \textbf{40.87} & 31.51 & \textbf{32.03} & 20.37 & 19.92 & \textbf{9.50} & 7.70 & 62.66 & 55.64 & 41.75 & 43.52 \\
MagNet \cite{chng2024mask} & \textit{CVPR 24} & 36.36 & 35.90 & 30.76 & 29.92 & 23.58 & 23.29 & 16.88 & 15.07 & 8.34 & 6.82 & 62.83 & 55.78 & 33.73 & 33.27 \\
FIANet \cite{lei2024exploring} & \textit{TGRS 25} & 7.18 & 6.04 & 3.83 & 3.87 & 1.64 & 1.36 & 0.41 & 0.26 & 0.00 & 0.03 & 11.56 & 13.28 & 10.74 & 8.29 \\
Aeroformer \cite{li2025aeroreformer}& \textit{Arxiv 25} & 44.36 & 42.50 & 36.91 & 34.89  & 27.89&25.57 & 17.70 & 16.08 & 8.00 & 6.47 & 61.86 & 56.77 & 40.61 & 39.27 \\

MRSNet & -- & \textbf{48.39} & 47.65 & \textbf{40.87} & 39.70 & \textbf{33.01} & 30.90 & \textbf{22.08} & \textbf{20.47} & 9.23 & \textbf{8.54} & \textbf{63.59} & \textbf{57.04} & \textbf{44.86} & \textbf{43.60} \\
\hline
\hline
\end{tabular}
\label{tab2}
\end{table*}

\section{Experiment}
\subsection{Dataset Setup and Evaluation Metrics}

To fully verify the effectiveness and generalization of our proposed model, we conduct experiments on both our self-constructed NWPU-Refer dataset and a public dataset. For NWPU-Refer, the dataset is divided into training, validation, and test sets using a 7:1:2 ratio based on the sample distribution across different regions. For RRSIS-D, we follow the dataset split and experimental protocol of the original paper to ensure fair comparison with prior methods. To comprehensively evaluate the performance of the model, the following metrics are selected: precision at thresholds (P@0.5, P@0.6, P@0.7, P@0.8, P@0.9), overall IoU (oIoU), and mean IoU (mIoU).

\subsection{Experimental Settings}
All experiments are performed using PyTorch and trained on an NVIDIA GTX 6000 GPU. The AdamW optimizer is employed with an initial learning rate of 6e-4, and a cosine annealing strategy is used to dynamically adjust the learning rate, facilitating faster convergence. To ensure fair comparisons, only the English language is used for training. MRSNet uses a Swin-Base visual encoder with patch size 4 and window size 12, together with a bert-base-uncased language encoder.

\begin{figure*}[ht]
\centering
\includegraphics[width=6.2in]{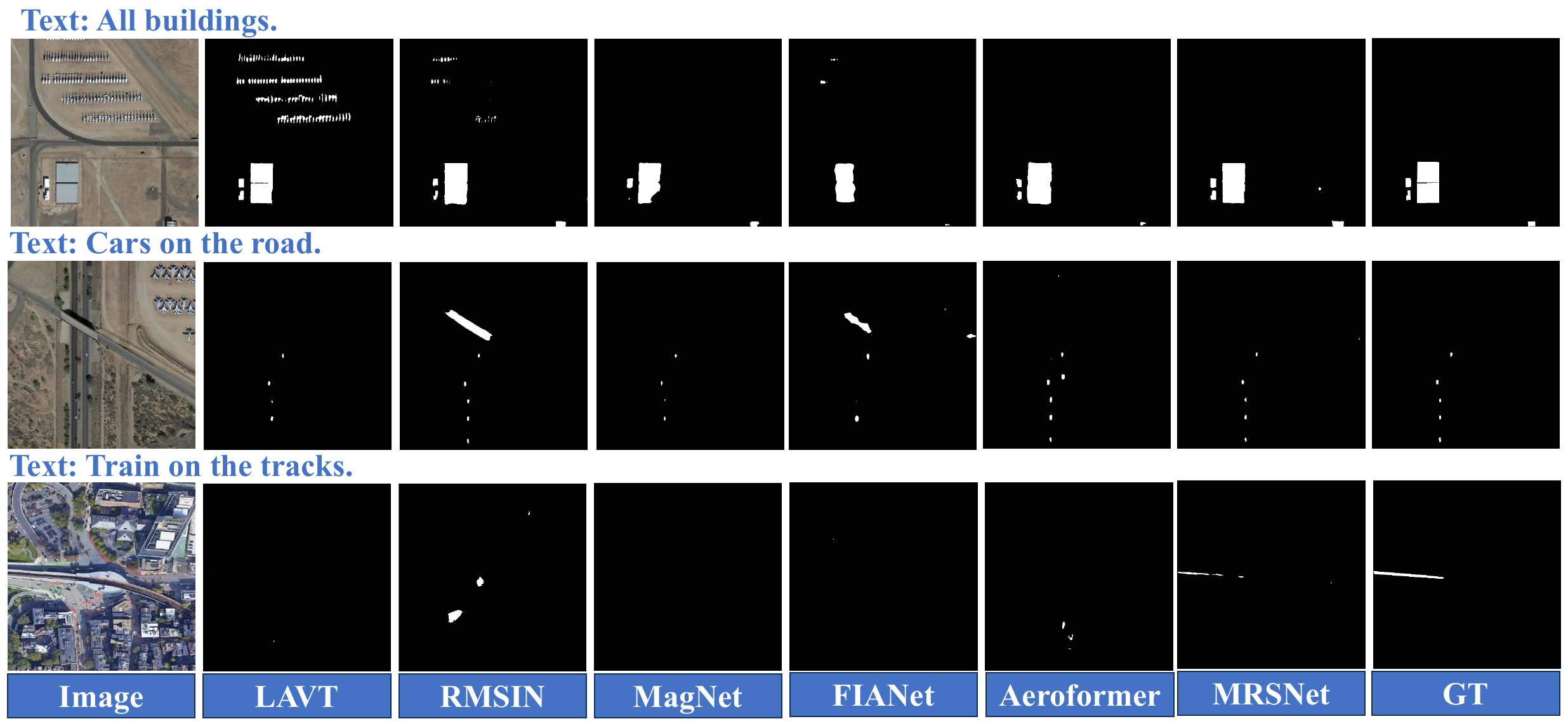}
\caption{Qualitative comparison on representative NWPU-Refer queries.}
\label{fig_8}
\end{figure*}

\subsection{Comparison with Existing Methods}

We compare MRSNet with representative baselines, including the general referring segmentation models LAVT and MagNet, the remote-sensing-oriented RMSIN and Aeroformer, and the feature-interaction model FIANet. Table \ref{tab2} reports the results on NWPU-Refer. MRSNet achieves the strongest overall performance on this benchmark, especially on the validation split and on strict-threshold metrics that emphasize accurate boundaries. On the test split, MRSNet surpasses RMSIN at P@0.8, P@0.9, oIoU, and mIoU, while RMSIN remains slightly better at P@0.5--P@0.7. This pattern indicates that MRSNet favors finer mask quality and spatial integrity rather than coarse localization alone. This behavior is consistent with the architectural design. IFIM improves local detail recovery and relation-aware filtering, which is particularly beneficial under stricter IoU thresholds, whereas HFIM preserves cross-scale consistency for IoU-based metrics. RMSIN retains an advantage on some lower-threshold scores, likely because its rotated decoding branch is specialized for coarse orientation-aware localization. In contrast, MRSNet prioritizes high-confidence mask quality, which is more directly reflected in P@0.8, P@0.9, oIoU, and mIoU.

\begin{table*}[ht]
\caption{Performance comparison across methods on the RRSIS-D dataset.}
\centering
\renewcommand{\arraystretch}{1.1}
\setlength{\tabcolsep}{4.0pt}

\begin{tabular}{l|c|c|c|c|c|c|c|c|c|c|c|c|c|c|c}
\hline
\hline
\multirow{2}{*}{\textbf{Method}} & \multirow{2}{*}{\textbf{Publication}} & \multicolumn{2}{c|}{\textbf{P@0.5}} & \multicolumn{2}{c|}{\textbf{P@0.6}} & \multicolumn{2}{c|}{\textbf{P@0.7}} & \multicolumn{2}{c|}{\textbf{P@0.8}} & \multicolumn{2}{c|}{\textbf{P@0.9}} & \multicolumn{2}{c|}{\textbf{oIoU}} & \multicolumn{2}{c}{\textbf{mIoU}} \\
\cline{3-16}
& & \textbf{val} & \textbf{test} & \textbf{val} & \textbf{test} & \textbf{val} & \textbf{test} & \textbf{val} & \textbf{test} & \textbf{val} & \textbf{test} & \textbf{val} & \textbf{test} & \textbf{val} & \textbf{test} \\
\hline
RRN \cite{li2018referring} & \textit{CVPR 18} & 51.09 & 51.07 & 42.47 & 42.11 & 33.04 & 32.77 & 20.80 & 21.57 & 6.14 & 6.37 & 66.53 & 66.43 & 46.06 & 45.64 \\
CMSA \cite{ye2019cross} & \textit{CVPR 19} & 55.68 & 55.32 & 48.04 & 46.45 & 38.27 & 37.43 & 26.55 & 25.39 & 9.02 & 8.15 & 69.68 & 69.39 & 48.85 & 48.54 \\
LSCM \cite{hui2020linguistic}& \textit{ECCV 20} & 57.12 & 56.02 & 48.04 & 46.25 & 37.87 & 37.70 & 26.37 & 25.28 & 7.93 & 8.27 & 69.28 & 69.05 & 50.36 & 49.92 \\
CMPC \cite{huang2020referring} & \textit{CVPR 20} & 57.93 & 55.83 & 48.85 & 47.40 & 38.50 & 36.94 & 25.28 & 25.45 & 9.31 & 9.19 & 70.15 & 69.22 & 50.41 & 49.24 \\
BRINet \cite{hu2020bi} & \textit{CVPR 20} & 58.79 & 56.90 & 49.54 & 48.77 & 39.65 & 39.12 & 28.21 & 27.03 & 9.19 & 8.73 & 70.73 & 69.88 & 51.14 & 49.65 \\
CMPC+ \cite{liu2021cross} & \textit{TPAMI 21} & 59.19 & 57.65 & 49.36 & 47.51 & 38.67 & 36.97 & 25.91 & 24.33 & 8.16 & 7.78 & 70.14 & 68.64 & 51.41 & 50.24 \\
LGCE \cite{yuan2024rrsis} & \textit{TGRS 24} & 68.10 & 67.65 & 60.52 & 61.53 & 52.24 & 51.45 & 42.24 & 39.62 & 23.85 & 23.33 & 76.68 & 76.34 & 60.16 & 59.37 \\
LAVT \cite{yang2022lavt} & \textit{CVPR 22} & 69.54 & 69.52 & 63.51 & 63.63 & 53.16 & 53.29 & \textbf{43.97} & \textbf{41.60} & \textbf{24.25} & \textbf{24.94} & \textbf{77.59} & \textbf{77.19} & 61.46 & 61.04 \\
GeoGround & \textit{ArXiv 25} & 68.69 & 67.50 & - & - & - & - & - & - & - & - & - & - & 61.10 & 60.50 \\
MRSNet & -- & \textbf{71.38} & \textbf{70.81} & \textbf{64.20} & \textbf{64.46} & \textbf{53.51} & \textbf{53.52} & 42.53 & 40.36 & 23.51 & 23.81 & 77.16 & 76.43 & \textbf{61.71} & \textbf{61.15} \\
\hline
\hline
\end{tabular}
\label{tab_rrsisd}
\end{table*}
To further assess the proposed method on another public dataset, we additionally conduct experiments on RRSIS-D. Table \ref{tab_rrsisd} reports the published baselines together with our MRSNet results on the validation and test splits. MRSNet achieves competitive performance in terms of both mIoU and oIoU, demonstrating competitive performance on this benchmark. Compared with the fully reported baselines, MRSNet obtains the best mIoU on both splits and also outperforms LGCE and LAVT on P@0.5--P@0.7, indicating stronger overall target localization. At the same time, LAVT remains slightly better on P@0.8--P@0.9 and oIoU, which suggests that there is still room to improve strict-threshold mask precision. We also introduce GeoGround, a recently proposed transformer-based method designed for remote sensing image grounding, for reference. Although GeoGround only reports partial test metrics in its original implementation, MRSNet still achieves clear improvements over its published test mIoU score. Overall, these results confirm that our MRSNet maintains favorable effectiveness when evaluated on this additional public RRSIS dataset.
\begin{figure*}[ht]
\centering
\includegraphics[width=7.2in]{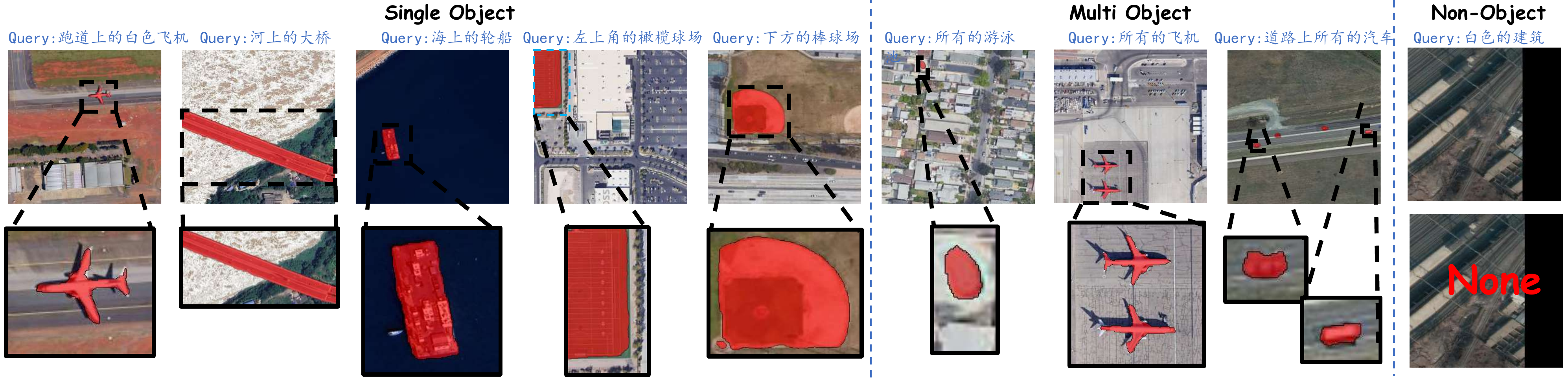}
\caption{Visualization results of MRSNet for Chinese referring segmentation. The left, middle, and right parts show single-object, multi-object, and non-object queries, respectively.}
\label{fig_7}
\end{figure*}
\subsection{Visual Results}
Qualitative results on NWPU-Refer are presented in Fig.~\ref{fig_8}, where MRSNet is compared with LAVT, RMSIN, MagNet, FIANet, and Aeroformer. The selected examples cover three representative yet challenging scenarios, namely building extraction, road-vehicle segmentation, and train-on-track localization, which involve dense distributions, small targets, and strict spatial positioning requirements. As can be seen, MRSNet accurately localizes the target regions under the guidance of the referring expressions and produces complete, high-quality segmentation masks. In the building extraction example, MRSNet effectively suppresses background interference and recovers a complete and continuous building region with clear boundaries, avoiding the fragmentation and incompleteness observed in the competing methods. In the road-vehicle scenario, MRSNet more precisely covers the target vehicles while preserving the shape and distribution of each instance, with fewer omissions, false detections, and adhesions to adjacent background regions. In the train localization example, MRSNet produces an accurate response along the railway direction, achieving better spatial alignment with the text description and ground truth while preserving the slender structure and overall integrity of the train. These visual comparisons further verify the effectiveness of the proposed intra-scale and cross-scale feature interaction design. By better capturing fine-grained spatial details and modeling target relationships in complex remote sensing scenes, MRSNet yields more complete target regions, more accurate localization, and higher-quality masks than the competing methods.

As illustrated in Fig.\ref{fig_7}, we show Chinese-query segmentation results on NWPU-Refer. The displayed queries are manually written to cover single-object, multi-object, and non-object settings. In the single-object cases, the model accurately identifies targets such as a white airplane on the runway, a bridge over a river, and a ship at sea. In the multi-object cases, it successfully segments all swimming pools, airplanes, and vehicles on the road despite substantial variation in size and distribution. The non-object cases further show that MRSNet can suppress irrelevant regions when the queried target does not exist. These examples highlight the robustness of the model under bilingual referring expressions and complex remote sensing scenes.

\subsection{Ablation Study}
We conduct ablation experiments on NWPU-Refer to verify the contributions of the three IFIM submodules, namely Pyramidal Spatial-Spectral Refinement (PSR), Context-Aware Spatial Relation Modeling (CSR), and Cross-Modal Alignment (CMA), together with the Hierarchical Feature Integration Module (HFIM). Table~\ref{tab3} summarizes the results in a unified module-wise manner. In particular, Model4 isolates the effect of HFIM by removing only the hierarchical cross-scale fusion stage while keeping the full IFIM branch unchanged. Specifically, when PSR or CSR is marked with ``$\times$'', the corresponding branch is completely removed from IFIM without introducing an auxiliary replacement path. In contrast, the CMA ablation replaces the original cross-modal alignment branch with a visual-only projection module of matched output dimensionality, and the HFIM ablation removes the hierarchical cross-scale fusion stage entirely and feeds the refined stage-wise features directly to the decoder.

\begin{table*}[ht]
\caption{Ablation results on the NWPU-Refer test set.}
\centering
\renewcommand{\arraystretch}{1.1}
\setlength{\tabcolsep}{4pt}

\begin{tabular}{l|c|c|c|c|c|c|c|c|c}
\hline
\hline
\multirow{2}{*}{\textbf{Setting}} & \multicolumn{3}{c|}{\textbf{IFIM}} & \multirow{2}{*}{\textbf{HFIM}} & \multirow{2}{*}{\textbf{P@0.7}} & \multirow{2}{*}{\textbf{P@0.8}} & \multirow{2}{*}{\textbf{P@0.9}} & \multirow{2}{*}{\textbf{oIoU}} & \multirow{2}{*}{\textbf{mIoU}} \\
\cline{2-4}
 & \textbf{PSR} & \textbf{CSR} & \textbf{CMA} &  &  &  &  &  &  \\
\hline
Model1 & $\checkmark$ & $\times$ & $\checkmark$ & $\checkmark$ & 21.57 & 12.51 & 4.35 & 48.04 & 36.82 \\
Model2 & $\times$ & $\checkmark$ & $\checkmark$ & $\checkmark$ & 23.18 & 16.21 & 6.16 & 52.10 & 40.08 \\
Model3 & $\checkmark$ & $\checkmark$ & $\times$ & $\checkmark$ & 14.98 & 9.45 & 3.54 & 31.64 & 25.21 \\
Model4 & $\checkmark$ & $\checkmark$ & $\checkmark$ & $\times$ & 27.06 & 17.02 & 6.63 & 55.91 & 41.22 \\
MRSNet & $\checkmark$ & $\checkmark$ & $\checkmark$ & $\checkmark$ & \textbf{30.90} & \textbf{20.47} & \textbf{8.54} & \textbf{57.04} & \textbf{43.60} \\
\hline
\hline
\end{tabular}

\label{tab3}
\end{table*}

Three observations can be drawn from Table~\ref{tab3}. First, the performance degradation of Model1 and Model2 compared to MRSNet demonstrates the mutual complementarity of PSR and CSR in intra-scale feature refinement. Specifically, CSR contributes more to performance improvement than PSR, as reflected by the higher mIoU of Model2 than Model1, indicating that context-aware spatial relation modeling plays a more critical role in capturing fine-grained spatial information.
Second, Model3 exhibits the most severe performance drop compared to MRSNet, with mIoU decreasing by 18.39\% and oIoU by 25.40\%. Model3 replaces CMA with a visual-only projection. This confirms that explicit cross-modal alignment is indispensable for bridging the semantic gap between visual and linguistic features, and pure visual refinement cannot substitute for cross-modal interaction in referential segmentation.
Third, Model4 shows a consistent performance decline relative to MRSNet, with mIoU decreasing from 43.60 to 41.22 and oIoU decreasing from 57.04 to 55.91. The degradation is more pronounced under stricter thresholds, where P@0.7, P@0.8, and P@0.9 drop by 3.84, 3.45, and 1.91 points, respectively. This indicates that HFIM is particularly beneficial for preserving multi-scale consistency and refining high-confidence masks, rather than merely improving coarse localization.

\section{Conclusion} 
In this paper, we present NWPU-Refer, a large-scale bilingual RRSIS dataset that offers clear advantages in scene diversity, annotation richness, and practical task coverage over existing benchmarks. We also present the Multi-scale Referring Segmentation Network, a sophisticated model designed to address the challenges of RRSIS. MRSNet effectively extracts object features and their interrelationships across different scales through the use of an Intra-scale Feature Interaction Module at each encoder stage. Furthermore, a Hierarchical Feature Interaction Module is designed to ensure comprehensive integration of global semantic information and local detailed features, enabling the model to handle complex spatial variations and intricate object relationships.
Experiments on NWPU-Refer and RRSIS-D show that MRSNet achieves competitive overall performance, setting a new benchmark for future research in this field.

\bibliographystyle{IEEEtran}
\bibliography{reference}

\end{document}